\documentclass[conference]{IEEEtran}
\IEEEoverridecommandlockouts

\usepackage{cite}
\usepackage{amsmath,amssymb,amsfonts}
\usepackage{algorithmic}
\usepackage{graphicx}
\usepackage{textcomp}
\usepackage{xcolor}

\usepackage{booktabs}
\usepackage[table, dvipsnames]{xcolor}
\usepackage{amssymb}
\usepackage{amsmath}
\usepackage[table]{xcolor}
\usepackage{soul}
\usepackage{color}   
\usepackage{hyperref}
\usepackage{subcaption}
\usepackage{caption}
\usepackage{float}
\usepackage{multirow}

\def\BibTeX{{\rm B\kern-.05em{\sc i\kern-.025em b}\kern-.08em
    T\kern-.1667em\lower.7ex\hbox{E}\kern-.125emX}}

\begin{document}

\title{MOSAIC: Multi-agent Orchestration for Task-Intelligent Scientific Coding
\thanks{}
}

\author{\IEEEauthorblockN{1\textsuperscript{st} Siddeshwar Raghavan$^{*}$\thanks{$^{*}$ Work was done as a Givens Associate at Argonne National Laboratory}}
\IEEEauthorblockA{\textit{Electrical and Computer Engineering} \\
\textit{Purdue University}\\
West Lafayette, IN, USA \\
raghav12@purdue.edu}
\and
\IEEEauthorblockN{2\textsuperscript{nd} Tanwi Mallick}
\IEEEauthorblockA{\textit{Mathematics and Computer Science Division} \\
\textit{Argonne National Laboratory}\\
Lemont, IL, USA \\
tmallick@anl.gov}
}

\maketitle

\begin{abstract}
We present MOSAIC, a multi-agent Large Language Model (LLM) framework for solving challenging scientific coding tasks. Unlike general-purpose coding, scientific workflows require algorithms that are rigorous, interconnected with deep domain knowledge, and incorporate domain-specific reasoning, as well as algorithm iteration without requiring I/O test cases. Many scientific problems also require a sequence of subproblems to be solved, leading to the final desired result. MOSAIC is designed as a training-free framework with specially designed agents to self-reflect, create the rationale, code, and debug within a student-teacher paradigm to address the challenges of scientific code generation. This design facilitates stepwise problem decomposition, targeted error correction, and, when combined with our Consolidated Context Window (CCW), mitigates LLM hallucinations when solving complex scientific tasks involving chained subproblems. We evaluate MOSAIC on scientific coding benchmarks and demonstrate that our specialized agentic framework outperforms existing approaches in terms of accuracy, robustness, and interpretability.
\end{abstract}

\begin{IEEEkeywords}
LLM for science, multi-agent framework, code generation
\end{IEEEkeywords}

\section{Introduction}
\label{sec:introduction}
Large Language Models (LLMs) have recently shown significant advancements, exhibiting capabilities in reasoning, strategic planning, and task execution that are comparable to human cognition. These developments pave the way for autonomous agents that can interpret objectives, devise strategies, and execute actions with minimal human intervention. Modern LLM architectures can decompose complex prompts, iteratively refine outputs, and collaborate with humans or other AI agents to solve intricate problems. These advancements have led to the development of multi-agent frameworks, where specialized agents coordinate to achieve goals that would be challenging for a single model.

Early code‐generation efforts explored a variety of techniques, including straightforward one‐shot synthesis~\cite{chen2021evaluatinglargelanguagemodels}, chain‐of‐thought prompting to break problems into reasoning steps~\cite{cot}, the use of generated test cases to guide output~\cite{chen2022codet}, retrieval‐augmented methods that incorporate external code examples~\cite{parvez2021retrieval}, and in‐context learning with curated exemplars~\cite{shum-etal-2023-automatic, zhang2022automatic}. More recent work has moved toward richer planning and search strategies constructing explicit multi‐step plans, employing sampling or tree‐search to explore candidate solutions~\cite{zhou2023language}, leveraging self‐retrieval for dynamic context expansion~\cite{yasunaga2023large}, and coordinating multiple specialized agents to tackle complex programming tasks~\cite{agentcoder, mapcoder,codesim, mallick2024chatvis}. These specialized multi-agent frameworks address general and competitive coding challenges by planning, executing, and iteratively debugging against simulated or provided sample I/O.

Despite these breakthroughs, scientific coding presents several unique challenges. \textbf{Scientific workflows require precise numerical methods and high reproducibility} where even minor deviations can undermine the results. Next is, \textbf{Domain specific reasoning} which requires embedding scientific concepts into the algorithms leading to the code generated. Moreover, scientific problems are often broken down into multiple interdependent sub problems, creating \textbf{long chains of reasoning where the errors can propagate}. Preserving context at this scale is also challenging, as sub problems are growing. LLMs struggle to retain critical information and become increasingly prone to \textbf{hallucinations and inconsistencies when operating near their context limits}~\cite{LLM_hallucinate}, reducing their reliability for end-to-end scientific code generation. Collectively, these requirements often exceed the current capabilities of both monolithic LLMs and simple agent setups. The need for intricate data dependencies and formal validation further complicates code generation and can introduce subtle, difficult-to-detect errors.

Another key difference is, scientific coding datasets provide function signatures but no I/O examples, and generating them is non-trivial as this is the algorithm we're trying to generate with our agentic framework! This fundamental difference makes prior multi-agent frameworks such as MapCoder~\cite{mapcoder}, CodeSIM~\cite{codesim}, and AgentCoder~\cite{agentcoder} unsuitable for scientific settings, as their verification and algorithm iteration loops rely on the availability of test cases. 

Motivated by this research gap, we introduce MOSAIC, a fully autonomous, LLM-agnostic multi-agent framework for scientific code generation that operates without the need for validation I/O test cases. At a high level, MOSAIC orchestrates specially designed agents to decompose problems, self-reflect on the algorithm, generate and refine code and maintain context across chained subproblems.

Our comprehensive evaluation across three LLM backbones and four optimization baselines (direct synthesis, chain-of-thought prompting, self-planning, analogical reasoning) demonstrates that MOSAIC substantially enhances scientific code generation, outperforming every comparative method. In particular, when paired with GPT-4o, MOSAIC achieves a 20.01\% main problem–solving rate and a 41.69\% subproblem–solving rate, a 8.5\% improvement over the baseline and a 24\% improvement over other methods.

To summarize, our main contributions presented are as follows -
\begin{enumerate}
    \item We introduce MOSAIC, a training free, LLM-agnostic multi agentic framework explicitly designed for scientific code generation in test I/O free environments. 
    \item We design and integrate four agents that collaboratively decompose scientific problems, self-reflect and formulate algorithms, manage context across chained subproblems and iteratively code and debug for improved accuracy and executability.
    \item Through extensive experiments on SciCode dataset, we show that MOSAIC consistently outperforms all baselines, achieving up to 24\% higher problem-solving accuracy.
    \item Finally, targeted ablation studies reveal which agents contribute most critically, highlighting the value of orchestrating distinct experts to tackle complex tasks.
\end{enumerate}

\begin{figure*}[ht]
    \centering
    \includegraphics[width=1\linewidth]{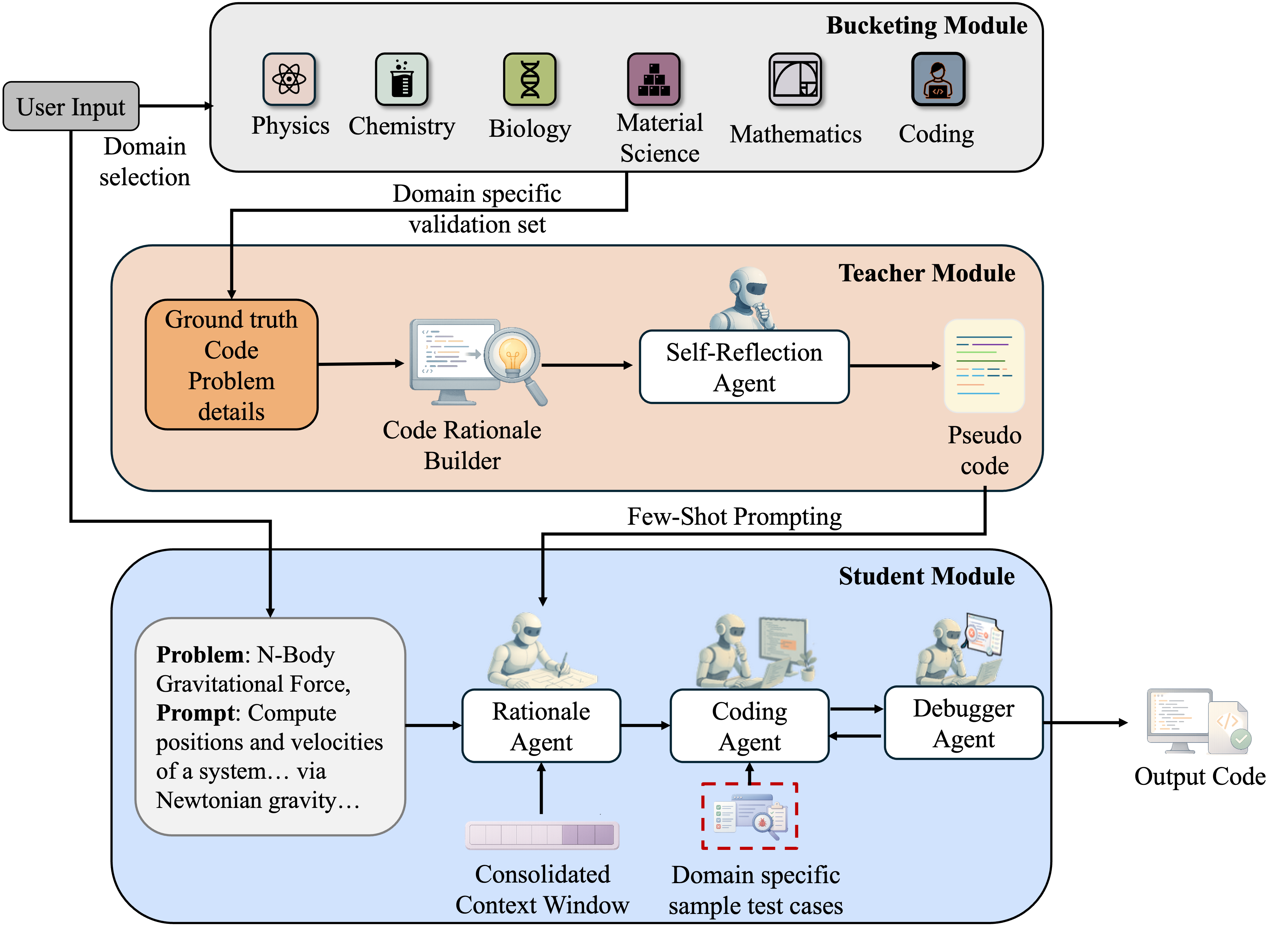}
    \caption{Overview of MOSAIC, a four-agent framework with domain independent memory. The design is inspired by knowledge distillation, where the Teacher module leverages few-shot examples from domain validation data to guide the Student module. This process enables the generation of clean rationales, which are then converted into accurate and executable code. The consolidated context window helps the agents focus on the current problem without being overwhelmed by previously generated information. (NOTE: When available, sample test cases from the dataset can also be incorporated by the Coding Agent)}
    \label{fig:mosaic}
\end{figure*}


\section{Related Work}
\label{sec:rel_work}
\subsection{LLM for Science}

In recent years, Scientific research workflows have begun to adopt the use of Large Language Models (LLMs). Unlike general-purpose dialogue assistants, the requirements of scientific LLMs depend on their capacity to retrieve, reason, and construct a hypothesis within the proposed domain. Recent LLM surveys have highlighted their potential to accelerate scientific discovery pipelines, reduce the time required for hypothesis generation, and improve the reproducibility of experiments across various domains~\cite{Wang2023ScientificDI, ai4science2023impactlargelanguagemodels, zhang2024scientificllms_survey} with human supervision. Models such as Galactica~\cite{taylor2022galactica} and ~BioGPT ~\cite{biogpt} showed that domain-specific pretraining significantly improves the ability to find literature and suggest valuable insights. More recent work has focused on multimodal scientific assistants, where textual reasoning is combined with images, graphs, or laboratory records \cite{edwards2024sciem, li2024matbenchllm}.

One of the primary challenges with LLMs is the reliability of reasoning across multiple steps. Techniques like Chain-of-Thought (CoT)~\cite{cot}, Self-Consistency~\cite{wang2023selfconsistencyimproveschainthought}, and Self-Refine~\cite{selfrefine} aim to strengthen this process by guiding models to produce clearer, more structured reasoning. Newer algorithms, such as Tree-of-Thoughts \cite{treeofthoughts} and Graph-of-Thoughts \cite{graphoffthoughts}, extend this by introducing search trees or graphs over reasoning paths, allowing science agents to remove unproductive solution strategies. Self-reflection and reinforcement learning methods, such as Reflexion~\cite{reflexion}, show particular promise in an iterative trial and error method that mimics the experimental cycle.

Alongside methodological advances, benchmarks have evolved to probe scientific reasoning. Biomedical datasets, such as PubMedQA ~\cite{jin2019pubmedqa} and MedQA~\cite{jin2021medqa}, benefit from real reasoning in medical literature. In contrast, ScienceQA~\cite{lu2022scienceqa} uses diagrams, text, and scientific concepts for multimodal reasoning.  Recent works focus on dependability, BenchFlow~\cite{rodriguez2023benchflow} evaluates reproducibility of generated protocols, while FACE4RAG~\cite{face4rag} focuses on factual consistency in retrieval-augmented systems. For executable code generation in science, SciCode~\cite{scicode} evaluates whether models can produce working programs across five domains(physics, chemistry, biology, mathematics, and material science) combining scientific reasoning with code executability. SciCode benchmark also exposes the limitations of existing methods in producing domain specific solutions that are not only relevant but also executable. This points to the need for new frameworks that can more effectively integrate reasoning, code generation, and verification into a unified process.

\subsection{LLM for code generation}

LLMs have also transformed code generation and software engineering workflows. Early works such as Codex~\cite{chen2021evaluating} showed that pretraining on source code databases and fine-tuning on problem solution pairs can improve code completions in LLMs. Newer works, including CodeGen~\cite{codegen}, SantaCoder~\cite{santacoder}, and StarCoder~\cite{starcoder}, have expanded to support multiple programming languages and multi-step problems. Benchmark suites such as HumanEval~\cite {chen2021evaluating} and APPS~\cite{APPS} focus on competitive level programming tasks for LLMs, and MBPP~\cite{MBPP} focuses on simpler Python coding problems to evaluate the accuracy of LLM generated code. However, most general coding datasets provide validation test cases to help LLM improve the algorithm generated before testing on unseen cases.

To move beyond toy benchmarks, SWE-bench~\cite{swebench} evaluates whether LLMs can resolve real issues in active GitHub repositories. Their findings reveal a considerable gap between solving isolated prompts and producing code snippets that are ready for integration. SWE-agent~\cite{sweagent} addresses part of this challenge by providing LLMs with access to repositories, compilers, and test suites, leading to improvements in performance.

\subsection{Agentic LLM frameworks}
A recent direction of research is agentic frameworks, where multiple instances of LLMs work as individual components within a collaborative framework. Each agent is assigned a specialized role, such as planning, coding, or verification. Within this framework, these individual agents interact with each other to tackle complex programming and reasoning tasks more effectively than a single LLM model. Recently, multi-agent frameworks have been proposed to imitate the human programming cycle. MapCoder~\cite{mapcoder} and CodeSim~\cite{codesim} introduces a pipeline where planning, coding, and debugging agents internally verify inputs and outputs before final code generation. These pipelines improve performance on HumanEval, MBPP and APPS benchmarks, suggesting that dividing roles among specialized agents helps reduce hallucination and error propagation. However, while effective in pure programming tasks, these systems do not directly generalize to scientific problem solving, where reasoning must be coupled with domain knowledge and robust debugging without the availability of validation Input/Output to improve algorithms generated.

Apart from specialized roles, some frameworks also enable the use of external APIs and existing tools~\cite{mathchat, toolformer, autogen} (ArXiV databases, mathematical toolkits like Wolfram Alpha). Open-source frameworks, such as LangGraph, put these ideas into practice by supporting planning and tool use within an organized workflow. These directions highlight the shift from “static” prediction/ generation systems to more autonomous and interactive systems for scientific discovery. Despite this progress, existing approaches remain limited in their ability to maintain coherent reasoning chains while generating executable code across diverse scientific domains without using external tools and API calls. Motivated by this challenge, we propose a modular framework designed to address these shortcomings for generating scientific code that is both executable and logically consistent. We follow an training free multi agentic approach by designing individual agents for self-reflection, rationale building, coding, and debugging. Our framework can break down complex scientific problems into solvable sub-tasks without any human supervision. In the following sections, we outline the design choices and core agents that form the foundation of our framework.


\section{Method}
\label{sec:method}
In this section, we present our framework, MOSAIC, detailing the various agents it orchestrates as well as the underlying design choices.

\subsection{MOSAIC Framework}
MOSAIC is a modular LLM-agnostic, multi-agent framework designed to break down complex scientific coding problems into accurate and executable solutions. Built on a unified and flexible architecture, MOSAIC can be customized for any scientific domain by embedding domain-specific knowledge into its agents without finetuning. The framework includes four primary agents: \textbf{Self-Reflection, Rationale, Coding}, and \textbf{Debugging}, which work together autonomously to generate and refine code. At the highest level, a \textbf{Bucketing Module} routes the problem to the appropriate scientific domain. While all domains use the same set of specialized agents, each maintains its own dedicated memory to prevent cross-domain interference.

Inspired by the concept of Knowledge Distillation (KD)~\cite{hinton2015distillingknowledgeneuralnetwork}, MOSAIC is structured as a student–teacher system, as illustrated in Figure~\ref{fig:mosaic}. Within the teacher component, we use ground-truth scientific code from a small subset ($\leq$ 5\%) of the training/validation data to create detailed rationales that break each problem into a sequence of solution steps. These step-by-step rationales, together with the corresponding ground-truth code, enable the teacher to guide the student in learning how to solve problems within a given scientific domain. In the following section, we explore the core agents that form the foundation of our framework.

\subsubsection{\textbf{Self-Reflection Agent}}

The agent receives the ground truth rationale and learns to evaluate its own intermediate reasoning steps. It identifies potential mistakes or omissions and refines its logic iteratively before arriving at the final pseudocode. By verbalizing its thought process and critically analyzing its reasoning path, the \textbf{Self-Reflection Agent} enhances output reliability, corrects logical flaws, and improves overall accuracy.

\subsubsection{\textbf{Rationale Agent}}
The \textbf{Rationale Agent} uses the generated pseudocode as few-shot examples to process scientific prompts from the test set. It then produces a clear, step-by-step reasoning plan similar to the structured guidance offered by the teacher model. Scientific problems typically involve a sequence of dependent sub-problems that need to be addressed in a specific order to reach the final solution. To address the risk of hallucination as the context window grows~\cite{LLM_hallucinate, banerjee2024llmshallucinateneedlive}, we implement a \textbf{Consolidated Context Window (CCW)} within the Rationale Agent, as illustrated in Figure~\ref{fig:mosaic}. The CCW helps the agent remain focused on the current task and determine the logical next steps in the reasoning process.  To ensure efficiency, the CCW contains only prior function signatures and brief one-sentence summaries, rather than full code history.

\subsubsection{\textbf{Coding Agent}}

The Coding Agent uses the detailed plan provided by the Rationale Agent to generate the corresponding code block, maintaining awareness of both the subproblem and the broader problem context. If the prompt includes sample test cases, the Coding Agent can incorporate them to improve robustness and ensure functionality for those scenarios (general purpose coding datasets have sample I/O).

\subsubsection{\textbf{Debugger Agent}}
The final core agent in the framework is the \textbf{Debugger Agent}, which executes the generated code and performs up to $k$ rounds of error correction in collaboration with the Coding Agent. This iterative process resolves syntax and import errors, ensuring the final output is executable. As scientific coding benchmarks such as SciCode~\cite{scicode} often lack sample input-output test cases, and generating them is highly complex. Thus, making it challenging to refine algorithmic logic on the fly. To address this challenge and improve the quality of generated algorithms, we guide the agents in a teacher-student setup to produce accurate and executable code.

\section{Experiments}
\label{sec:exps}

In this section, we detail our experimental setup, including the datasets selected and the LLM backbones integrated into our MOSAIC framework. We assess MOSAIC across scientific, general purpose, and competitive coding benchmarks, comparing its results against both baseline approaches and current best performing methods. Finally, we conduct an extensive ablation study to highlight each agent’s individual contribution and provide detailed insights on the performance.

\subsection{Datasets and Comparison Approaches}
We evaluate MOSAIC on the challenging SciCode dataset~\cite{scicode}, which covers five domains (Physics, Chemistry, Biology, Mathematics, and Materials Science) and comprises 65 main problems split into 283 subproblems. Each subproblem provides a prompt, background context, a function signature for code generation without access to the gold standard ground truth code. The dataset provides a validation set of 15 main problems and 50 subproblems with ground truth code that is disjoint with the test set. We compare MOSAIC against four baseline methods: Direct, Chain of Thought (CoT), Self Planning, and Analogical. Because current state-of-the-art LLM coding frameworks rely on sample test cases to generate the code and do not support integrating multiple subproblems into a unified workflow, they cannot be evaluated directly on SciCode. The test suite provided by SciCode~\cite{scicode} evaluates the correctness of a problem if all the subproblems are executable and match with the ground truth target value. 
To demonstrate MOSAIC’s versatility, we also evaluate it alongside leading multi-agent coding frameworks on the general-purpose MBPP dataset~\cite{MBPP} (1,000 problems) and the HumanEval dataset~\cite{humaneval} (164 problems), as well as on the APPS benchmark~\cite{APPS}, which comprises over 5,000 problems spanning introductory, interview, and competition-level challenges. On the general purpose coding datasets, we compare MOSAIC against MapCoder~\cite{mapcoder} and CodeSIM~\cite{codesim} where it achieves comparable performance.

\subsection{Implementation Details}

We build on the open-source PyTorch implementation of SciCode~\cite{scicode} for our experiments. Within MOSAIC, we employ LangGraph to orchestrate agent communication and ensure reproducibility. For each problem domain, we have a dedicated memory for the agentic framework to ensure encountering only the domain knowledge and prevent cross-domain interference. Each agent (Self-Reflection, Rationale, Coding, and Debugging) is guided by tailored prompts that constrain it to its specific role and yield the outputs needed to arrive at the final solution (Included in the Appendix~\ref{sec:appendix}). In MOSAIC’s teacher module, designed for knowledge distillation via few-shot prompting, we sample twenty problems at random from the APPS training set (seed 1993)~\cite{APPS}, use the ten MBPP problems provided for few-shot examples~\cite{MBPP}, and include five problems from the HumanEval dataset~\cite{humaneval} taken out of the entire dataset.

We evaluate our MOSAIC framework, other baselines and benchmarks using Open AI GPT-4o, Claude Sonnet 4 and Gemini 2.5 Flash. In our ablation studies~\ref{sec:ablation} we explore other open source LLM backbones. 

\subsection{Evaluation Metrics}

In this paper, we adopt the SciCode evaluation protocol~\cite{scicode} for the scientific dataset, counting solved sub-problems and main problems. We also report metrics that quantify how closely our outputs align with the reference solutions in SciCode in our Ablation section~\ref{sec:ablation}, even for cases that don’t fully succeed, since precision and accuracy are critical in scientific problem solving. For MBPP~\cite{MBPP}, HumanEval~\cite{humaneval}, and APPS~\cite{APPS}, we report performance as the percentage of test cases passed.

\section{Results and Discussion}
\label{sec:results}

\begin{table*}[ht!]
\centering
\resizebox{\textwidth}{!}{%
\begin{tabular}{l|l|cc|cc|cc|cc|cc|cc}
\toprule
LLM Backbone & Methods 
  & \multicolumn{2}{c|}{Total} 
  & \multicolumn{2}{c|}{Physics} 
  & \multicolumn{2}{c|}{Chemistry} 
  & \multicolumn{2}{c|}{Biology} 
  & \multicolumn{2}{c|}{Material Science} 
  & \multicolumn{2}{c}{Mathematics} \\
\midrule
\multirow{6}{*}{GPT-4o}   
  & SciCode Baseline &  7/65 &  94/283  &   3/30 &  48/145 &   1/7 &   13/42 &   0/7 &   5/25 &   2/11&   24/50 &   1/10 &   4/24 \\
\cmidrule(l){2-14}
  & Analogical       &   1/65 &  32/283  &   1/30 &  18/145 &   0/7 &   2/42 &   0/7 &   3/25 &   0/11&   6/50 &   0/10 &   3/24 \\
\cmidrule(l){2-14}
  & CoT       &   2/65 &  38/283  &   1/30 &  21/145 &   0/7 &   2/42 &   0/7 &   3/25 &   1/11&   8/50 &   0/10 &   4/24 \\
\cmidrule(l){2-14}
  & LATS             &   4/65 &  49/283  &   2/30 &  34/145 &   0/7 &   2/42 &   0/7 &   3/25 &   2/11&   8/50 &   0/10 &   2/24 \\
\cmidrule(l){2-14}
  & MOSAIC (\textit{ours})
                    &  \textcolor{Cerulean}{\textbf{12/65}} & \textcolor{Cerulean}{\textbf{113/283}} &  \textcolor{Cerulean}{\textbf{4/30}} &  \textcolor{Cerulean}{\textbf{56/145}} &   \textcolor{Cerulean}{\textbf{2/7}} &  \textcolor{Cerulean}{\textbf{14/42}} &   \textcolor{Cerulean}{\textbf{0/7}} &   \textcolor{Cerulean}{\textbf{7/25}} &   \textcolor{Cerulean}{\textbf{3/11}} &  \textcolor{Cerulean}{\textbf{26/50}} &   \textcolor{Cerulean}{\textbf{3/10}} &   \textcolor{Cerulean}{\textbf{10/24}} \\
\midrule\midrule
\multirow{2}{*}{Claude Sonnet 4} 
  & SciCode Baseline &   9/65 &  109/283  &   4/30 &  71/145 &   1/7 &   13/42 &   1/7 &   9/25 &   2/11&   8/50 &   1/10 &   8/24 \\
\cmidrule(l){2-14}
  & MOSAIC (\textit{ours}) 
                    &   \textbf{\textcolor{Cerulean}{13/65}} &  \textbf{\textcolor{Cerulean}{118/283}}  &   \textbf{\textcolor{Cerulean}{4/30}} &  \textbf{\textcolor{Cerulean}{77/145}} &   \textbf{\textcolor{Cerulean}{2/7}} &   \textbf{\textcolor{Cerulean}{17/42}} &   \textbf{\textcolor{Cerulean}{1/7}} &   \textbf{\textcolor{Cerulean}{8/25}} &   \textbf{\textcolor{Cerulean}{3/11} }&  \textbf{ \textcolor{Cerulean}{8/50}} &   \textbf{\textcolor{Cerulean}{3/10}} &   \textbf{\textcolor{Cerulean}{8/24}} \\ \midrule \midrule
\multirow{2}{*}{Gemini 2.5 flash} 
  & SciCode Baseline &   7/65 &  112/283  &   5/30 &  67/145 &   1/7 &   14/42 &   1/7 &   11/25 &   2/11&   9/50 &   1/10 &   1/24 \\
\cmidrule(l){2-14}
  & MOSAIC (\textit{ours}) 
                    &   \textbf{\textcolor{Cerulean}{11/65}} &  \textbf{\textcolor{Cerulean}{117/283}}  &   \textbf{\textcolor{Cerulean}{5/30}} &  \textbf{\textcolor{Cerulean}{88/145}} &   \textbf{\textcolor{Cerulean}{2/7}} &   \textbf{\textcolor{Cerulean}{6/42}} &   \textbf{\textcolor{Cerulean}{1/7}} &   \textbf{\textcolor{Cerulean}{5/25}} &   \textbf{\textcolor{Cerulean}{2/11}} &   \textbf{\textcolor{Cerulean}{6/50}} &   \textbf{\textcolor{Cerulean}{1/10}} &   \textbf{\textcolor{Cerulean}{12/24}} \\
                     
\bottomrule
\end{tabular}%
}
\caption{Performance comparison between baselines and MOSAIC on scientific datasets with different LLM backbones. \textcolor{Cerulean}{\textbf{Best}} results are highlighted. The SciCode benchmark consists of 65 main problems comprising a total of 283 subproblems spanning physics, chemistry, biology, materials science, and mathematics. A problem is considered solved only when all of its subproblems pass the corresponding test suites.}
\label{tab:main_comp}
\end{table*}

\begin{figure*}[h]
    \centering
    \includegraphics[width=1\textwidth]{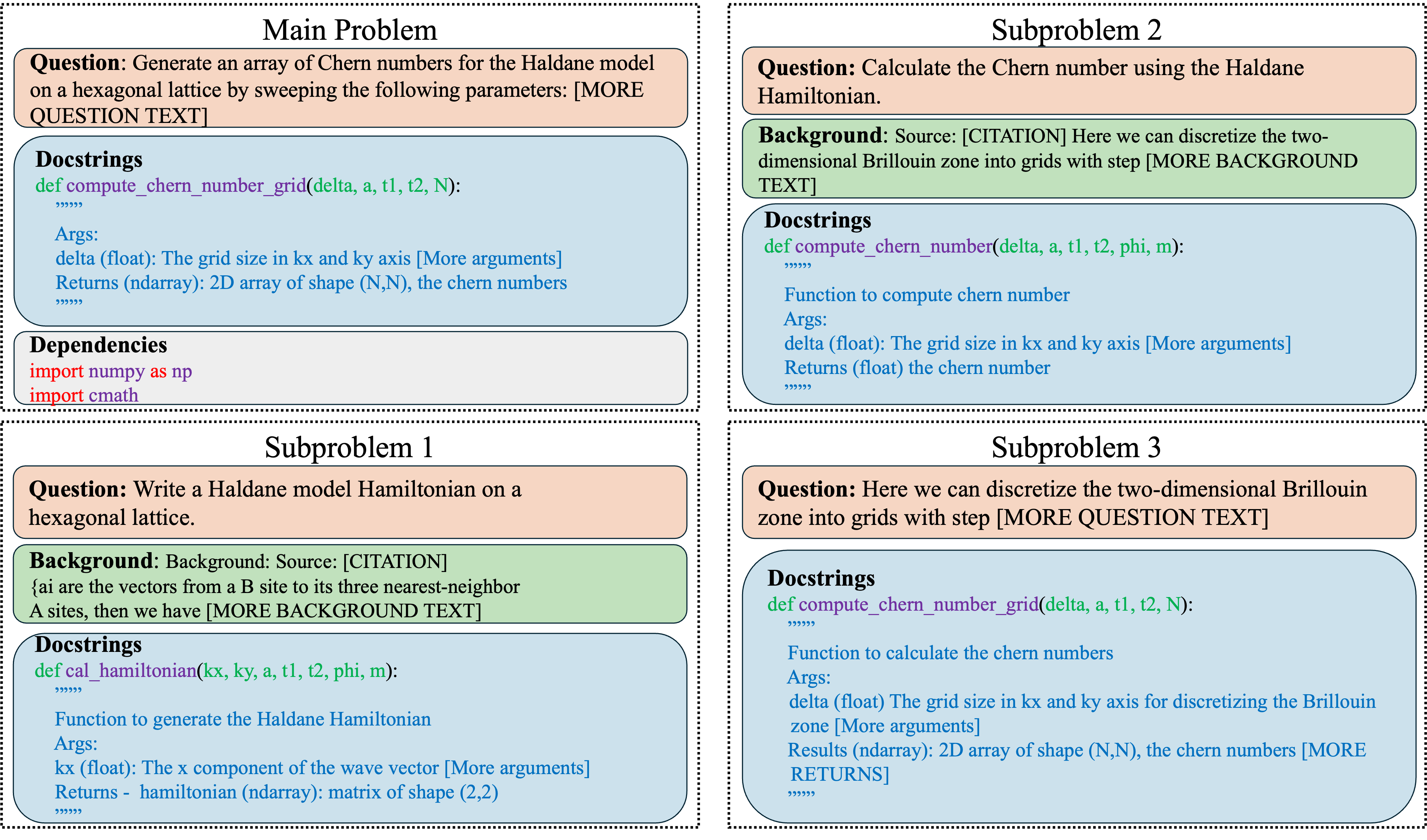}
    \caption{Structure of problems and subproblems in the SciCode dataset. Each main problem is composed of multiple subproblems, all of which must be solved correctly for the main problem to be considered successfully solved.}
    \label{fig:scicode_ds_struct}
\end{figure*}

In this section, we discuss the main findings of the paper and dive deeper into understanding the performance obtained. Table~\ref{tab:main_comp} presents the performance of different methods across multiple scientific domains using three LLM backbones: GPT-4o, Claude Sonnet 4, and Gemini 2.5 Flash. The evaluation is reported as the number of correct solutions out of the total attempted, with results broken down by subject areas including Physics, Chemistry, Biology, Material Science, and Mathematics. 
For GPT-4o, MOSAIC significantly outperforms other approaches, achieving (12/65)  correct problems and (113/283) correct subproblems across domains, with notable gains in Physics (56/145) and Material Science (26/50). Compared to baselines such as Analogical, CoT, and LATS, MOSAIC consistently demonstrates higher accuracy.
On Claude Sonnet 4, MOSAIC reaches the best performance with (13/65) total and (118/283) domain-specific correct answers, showing clear improvements in Physics (77/145) and Chemistry (17/42).
Similarly, for Gemini 2.5 Flash, MOSAIC achieves (11/65) overall and 117/283 across domains, again surpassing the SciCode baseline, with its strongest results in Physics (88/145) and Mathematics (12/24).
Overall, MOSAIC demonstrates robust and consistent improvements across all backbones and subject areas, highlighting the effectiveness of multi-agent orchestration in enhancing scientific problem-solving.
As shown in Table \ref{tab:main_comp}, we attribute MOSAIC’s gains over single-agent methods to its multi-agent orchestration strategy, inspired by knowledge distillation, which guides both algorithmic planning and execution. Unlike single-agent approaches such as Analogical, CoT, and LATS, MOSAIC leverages specialized agents for planning, code generation, and debugging, which reduces error propagation, mitigates hallucinations, and takes advantage of complementary strengths. This divide and conquer approach enables more consistent improvements across scientific domains and LLM backbones, resulting in higher accuracy and robustness.
The Rationale Agent proposes an initial plan, which the Self Reflection Agent refines via few-shot prompting with ground truth domain examples. Additionally, the Consolidated Context Window (CCW) helps maintain context by providing only the previously generated function signatures along with a one line summary of the implementation rather than the entire code and prompt information, and thereby helps to reduce hallucination and improve performance. The refined plan is then given to the Coding Agent which generates the code, and the Debugging Agent concentrates on syntactic corrections. Semantic validation remains limited because SciCode provides no means to validate the algorithmic correctness prior to evaluation. And generating reliable tests for scientific coding tasks is challenging as this is the algorithm the agentic framework is tasked with. As shown in Figure~\ref{fig:scicode_ds_struct}, the SciCode benchmark demands the use of context from previous subproblems and the main problem when addressing the current subproblem, given the inter dependence among them.

Table~\ref{tab:main_comp} shows that performance in the Biology domain is consistently lower than in the other domains, making it particularly challenging. During our experimentation, we observed incorrect order of steps and oversimplified algorithm logic. Such errors are qualitatively different from those in physics or mathematics, where failures are more toward numerical precision rather than conceptual. Addressing these domain-specific reasoning errors will likely require integrating structured biological knowledge rather than relying solely on statistical learning.
We identify three more contributing factors. First, current LLMs struggle to abstract and transfer concepts across related prompts, limiting generalization~\cite{xu2024llmsabstractionreasoningcorpus}. Second, the breadth and reliability of biological knowledge embedded in closed-source models are difficult to verify. Third, in the SciCode validation set, Biology is the least represented domain, leaving fewer ground-truth exemplars to teach the model through few-shot prompting, leaving many Biology problems unsolved.

We also observe an expected behavior pattern in very long problems (with more than 10 subproblems, sample structure depicted in Figure~\ref{fig:scicode_ds_struct}). The model often fails to maintain the necessary context, even with our Consolidated Context Window (CCW). In particular, it does not consistently reuse previously introduced function headers for inter-task connections or carry forward intermediate results into later subproblems. The combination of growing context, reduced supervision, and specific knowledge requirements makes this a challenging problem even with multi-agent orchestrations.

Despite these challenges, \textbf{MOSAIC} achieves substantial gains through a multi-agent design without any domain specific fine tuning. A primary contributor is the use of separate memory for each domain, which prevents interference across domains. With domain specific memory, the Rationale Agent produces more robust and contextually appropriate plans guided by the teacher module with few shot examples from the domain. For example, an effective strategy for a physics problem can differ significantly from that for a biology problem.

\begin{figure*}[!h]
    \centering
    \includegraphics[width=1\textwidth]{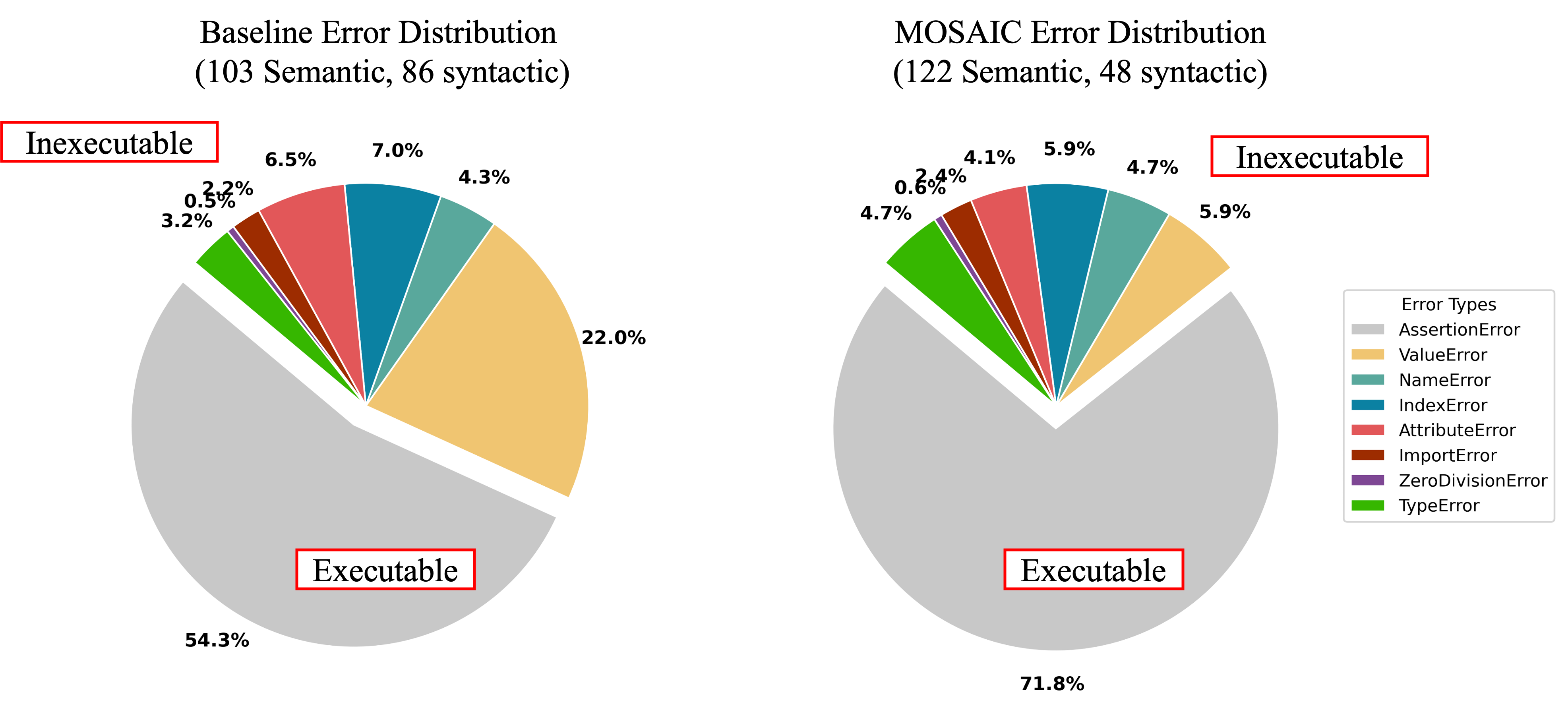}
    \caption{Error statistics on SciCode benchmarks. The figure distinguishes \textbf{syntactic errors} (failed execution) from \textbf{semantic errors} (output–target mismatches). Semantic errors are shown in gray, while other colors represent different categories of syntactic errors. MOSAIC substantially reduces both the overall error rate and the relative proportion of syntactic errors compared to the baseline.}
    \label{fig:error_types}
\end{figure*}
\subsection{Insights from different LLM backbones}
\label{subsec:llm_backbone}
From Table~\ref{tab:main_comp}, we notice that different LLM backbones exhibit slightly varying performance. A key understanding from these experiments is that differences in (closed) training make specific models excel in particular roles. For example, Claude Sonnet performs better at coding tasks than GPT-4o, while Gemini provides stronger reasoning capabilities compared to Claude. These observations highlight the potential of designing heterogeneous LLM backbones for agents within the same framework, which could further enhance overall performance. We deliberately chose to keep this aside for future work to concentrate on homogeneous model capabilities in a multi-agentic setup. 

Another insight we gathered from experimenting with various agents in MOSAIC is that LLMs sometimes ignore instructions issued by the orchestrator through prompt templates. Even small changes in prompts can lead to different outputs and in a chain of subproblems that error propagates fast, which needs to be mitigated. In practice, emphasizing critical instructions through capitalization proved effective in keeping the LLMs within the intended bounds and generating the expected outputs. From a theoretical perspective, this difference can be explained by tokenization, as “text” and “TEXT” are treated as distinct tokens. And by analogy to sentiment analysis, where capitalization suggests emphasis on the token.

\subsection{Errors in Output code}
Errors in written code can be broadly categorized into two types: \textbf{syntactic errors}, which prevent execution, and \textbf{semantic errors}, which refers to the logic of the algorithm being incorrect, leading to successful execution but a mismatch in target and output. In the context of code generation, ensuring executability is a primary prerequisite, as downstream correctness can only be evaluated once the code runs. Accordingly, MOSAIC prioritizes eliminating syntactic defects, even if the resulting outputs may not yet achieve full numerical precision.

Figure~\ref{fig:error_types} reports the distribution of error types across our benchmarks, offering a key perspective on MOSAIC’s improvements. With the exception of \texttt{AssertionError}, which arises from mismatches detected by the test suite, all other errors (\texttt{ValueError}, \texttt{TypeError}, \texttt{NameError}, \texttt{IndexError}, \texttt{AttributeError}, \texttt{ImportError}, \texttt{ZeroDivisionError}) are classified as syntactic, since they prevent successful execution. In contrast, \texttt{AssertionError} represents semantic errors, where code executes successfully but produces an incorrect output. The code from the baseline method results in nearly half of its generated programs failing due to syntactic errors, making them inexecutable and providing no algorithmic evidence of the failures. At first glance, MOSAIC appears to yield a larger percentage of errors. However, this shift is toward semantic errors rather than syntactic errors, and with a lower overall error count. This transition is preferred, as it allows us to focus on refining algorithmic rigor in the agents once execution is ensured, rather than repeatedly struggling to produce runnable code.

The domain segregated agentic design of MOSAIC plays a central role in achieving higher solve rates. The Debugger Agent systematically resolves execution errors, while the Self-Reflection and Rationale Agents improve logical consistency guided by the teacher module through in-domain examples, thereby reducing the likelihood of increasing syntactic failures.

\begin{table*}[h!]
\centering
\resizebox{0.7\textwidth}{!}{%
\begin{tabular}{c|c|c|c}
\hline
\multirow{2}{*}{Method} &
\multirow{2}{*}{\begin{tabular}[c]{@{}c@{}}HumanEval\\ (30 val / 134 test)\end{tabular}} &
\multirow{2}{*}{\begin{tabular}[c]{@{}c@{}}MBPP\\ (10 val / 500 test)\end{tabular}} &
\multicolumn{1}{c}{APPS} \\ 
 &  &  &  \\ \hline
Direct         & 89.63              & 48.80              & 12.70 \\ \hline
CoT            & 87.20              & 54.92              & 11.30 \\ \hline
Self-Planning  & 89.63              & 49.64              & 14.70 \\ \hline
Analogical     & 90.85              & 49.81              & 12.00 \\ \hline
MapCoder       & 90.26              & 77.92              & 22.37 \\ \hline
CodeSIM        & \textcolor{Cerulean}{\textbf{93.60}} & \textcolor{Bittersweet}{\textbf{80.49}} & \textcolor{Bittersweet}{\textbf{22.56}} \\ \hline
MOSAIC (\textit{ours}) & \textcolor{Bittersweet}{\textbf{92.53}} & \textcolor{Cerulean}{\textbf{84.90}} & \textcolor{Cerulean}{\textbf{24.71}} \\ \hline
\end{tabular}%
}
\caption{Performance comparison of different code generation methods on HumanEval, MBPP, and APPS benchmarks. Reported values represent accuracy scores (\%). The \textcolor{Cerulean}{\textbf{Best}} results for each benchmark are highlighted in blue, while the \textcolor{Bittersweet}{\textbf{second best}} results are highlighted in orange. MOSAIC achieves competitive performance with CodeSIM, ranking first on MBPP and APPS and second on HumanEval}
\label{tab:gen_pur_comp}
\end{table*}

\subsection{Performance on general coding benchmarks}

We evaluated MOSAIC on three popular general purpose and competitive coding benchmarks to compare its performance against existing LLM-based coding frameworks. As shown in Table~\ref{tab:gen_pur_comp}, MOSAIC outperforms the existing best methods on MBPP and APPS datasets and achieves comparable results on the HumanEval dataset. Although our framework is not designed for general purpose coding problems, its structured framework enables the generation of logical code leading to the observed performance numbers.

%
\section{Ablation Studies}
\label{sec:ablation}
In this section, we dive deep into the key components of our MOSAIC framework, analyze the performance on different LLM backbones and provide a comprehensive analysis on the performance numbers and pitfall regions where LLMs struggle to overcome challenges.

\begin{table*}[!th]
\resizebox{1\textwidth}{!}{%
\begin{tabular}{l|c|c|c|c|c|c}
\hline
Method           & Total        & Phys prob    & Chem     & Biology & Mat Sci & Math     \\ \hline
Baseline & 7/65, 94/283 & 3/30, 48/145 & 1/7, 13/42 & 0/7, 5/25 & 2/11, 24/50 & 1/10, 4/24 \\ \midrule
\begin{tabular}[c]{@{}l@{}}Baseline + Rationale +\\ Coding and Debug Agent\end{tabular} &
  9/65, 97/283 &
  3/30, 47/145 &
  2/7, 15/42 &
  0/7, 6/25 &
  3/11, 25/50 &
  1/10, 4/24 \\ \midrule
\begin{tabular}[c]{@{}l@{}}Baseline + Few-shot prompting + Rationale +\\ CCW(all prev. code) + Coding and Debug Agent\end{tabular} &
    4/65, 57/283 &
  1/30, 32/145 &
  1/7, 6/42 &
  0/7, 6/25 &
  1/11, 9/50 &
  1/10, 4/24 \\ \midrule
  \begin{tabular}[c]{@{}l@{}}Baseline + Self Reflection (stepwise) \\ + Few-shot prompting + Rationale \\+ CCW(prev. headers) + Coding and Debug Agent\end{tabular} &
  {6/ 65, 81/283} &
  {2/30, 48/145} &
  {1/7,  8/42} &
  {0/7, 6/25} &
  {2/11, 14/50} &
  {0/10, 5/24} \\ \midrule
  \begin{tabular}[c]{@{}l@{}}\textbf{MOSAIC(\textit{ours})} = \\ Baseline + Self Reflection (whole) \\ + Few-shot prompting + Rationale \\+ CCW(prev. headers) + Coding and Debug Agent\end{tabular} &
  \textbf{12/ 65, 113/283} &
  \textbf{4/30, 56/145} &
  \textbf{2/7,  14/42} &
  \textbf{0/7, 7/25} &
  \textbf{3/11, 26/50} &
  \textbf{3/10, 10/24} \\ \bottomrule
\end{tabular}
}
\caption{Performance comparison of MOSAIC with incremental addition of specialized agents (Rationale, Coding, Debugger) and mechanisms (Consolidated Context Window (CCW), Self-Reflection, and Few-shot prompting). The benchmark baseline~\cite{scicode} is included for reference. The results highlight how each component contributes to overall performance, with the full MOSAIC framework yielding the most significant improvements.}
\label{tab:abl_parts_comp}
\end{table*}

\subsection{Understanding Performance Gains of Different Components of MOSAIC}
The \textbf{MOSAIC} framework consists of four specialized agents designed to solve complex scientific coding problems. To understand the contribution of each component, we perform an ablation study that incrementally introduces different agents and mechanisms, reporting results in Table~\ref{tab:abl_parts_comp}. Since MOSAIC is built on top of the SciCode~\cite{scicode} framework, we use SciCode as our baseline which solves (7/65) main problems and (94/283) subproblems, and add MOSAIC agents step by step. For these experiments, we use OpenAI GPT-4o as the LLM model.

We first observe that adding a \textbf{Rationale Agent} together with iterative \textbf{Coding} and \textbf{Debugger Agents} improves performance to (9/65) main problems and (97/283) subproblems. This gain can be attributed primarily to a reduction in syntactic errors, leading to a 4\% increase in overall solve rate.

Next, we incorporate a \textbf{few-shot prompting} mechanism, which provides the ground-truth rationales from the validation problems together with the previously generated subproblem code in the context window, used alongside the \textbf{Coding} and \textbf{Debugger agents}. Interestingly, the combination of CCW (containing all prior code) and few-shot prompting, refer to Table~\ref{tab:abl_parts_comp} (line 3), resulted in a performance drop to (4/65) main problems and (57/283) subproblems, a drop by almost 4\% main problem solve rate before the introduction of self-reflection and consolidation strategies. Closer inspection of failure cases shows that the expanded context frequently led the model to replicate irrelevant fragments of earlier code, producing logically inconsistent or overly verbose solutions. In other words, an unrestricted CCW preserved excessive prior detail, reducing the model’s focus to the problem at hand. By contrast, restricting the CCW to earlier function signatures and concise one-line summaries improved performance but still lacked an abstraction mechanism to generalize reasoning across problems in a particular scientific domain.


\begin{table*}[t!]
\centering
\begin{tabular}{@{}l|l|l|l|cc}
\toprule
Method & Type          & LLM backbone     & Parameters & \multicolumn{2}{c}{Problems solved}      \\ \cmidrule(l){5-6} 
       &               &                  &            & \multicolumn{1}{c|}{Main} & Sub Problems \\ \midrule
\multirow{8}{*}{MOSAIC(\textit{ours})}       & Open Source   & Mistral          & 7B         & \multicolumn{1}{l|}{0/65}     & 24/283              \\
       & Open Source   & Gemma 3          & 27B        & \multicolumn{1}{l|}{1/65}     &  39/283              \\
       & Open Source   & Llama 4          & 16$\times$17B       & \multicolumn{1}{l|}{2/65}     & 41/283             \\
       & Open Source   & DeepSeek R1      & 32B       & \multicolumn{1}{l|}{4/65}     & 84/283              \\ \cmidrule(l){2-6} 
       & Closed Source & Gemini 2.5 Flash & NA         & \multicolumn{1}{l|}{11/65}     & 117/283              \\
       & Closed Source & Claudde Sonnet 4 & NA         & \multicolumn{1}{l|}{13/65}     & 118/283             \\
       & Closed Source & GPT-4o           & NA         & \multicolumn{1}{l|}{12/65}     & 113/283             \\ \bottomrule
\end{tabular}
\caption{Performance comparison of Open and Closed Source models on SciCode benchmark dataset. \textbf{Open-source models solve at most 4/65 main problems and 84/283 subproblems} (DeepSeek R1), while smaller ones like Mistral fail to solve any main problems. \textbf{Closed-source models perform substantially better}, with Claude Sonnet 4 achieving 13/65 main and 118/283 subproblems solved. On average, \textbf{closed-source} backbones solve nearly \textbf{3× more main problems} and \textbf{2× more subproblems}, highlighting the current performance gap.}
\label{tab:abl_cl_op_model}
\end{table*}

To address this, we introduce a \textbf{Self-Reflection Agent} that processes the ground-truth rationale and generates pseudocode. This pseudocode serves as a transferable abstraction that can generalize to other problems within the same scientific domain via few-shot prompting. We first experiment with \textbf{step-wise self-reflection}, where the rationale is processed one step at a time leading to (6/65) main problems and (81/283) subproblems solved. This approach underperforms compared to the baseline, as the overall problem context is fragmented and lost. However, when we instead apply \textbf{self-reflection over the entire rationale}, the context is preserved. This enables effective transfer of problem-solving knowledge across related tasks, resulting in (12/65) main problems and (113/283) subproblems solved, an 8.5\% performance gain over the baseline.

These ablation studies reveal two critical insights.\textbf{First, adding agents/  mechanisms in isolation led to reduced performance} as seen with unrestricted CCW and stepwise self-reflection, which introduced noise by enlarging the context window and losing context with individual steps. The results show that careful orchestration is vital, simply stacking components does not guarantee improvement. \textbf{Second, despite these intermediate drops, the agents are complementary} with each configuration solving a slightly different subset of problems across domains. Because the overlap among solved problems is minimal, integrating all agents and strategies in MOSAIC ultimately broadens coverage and yields significantly higher overall performance than the baseline.

\subsection{Comparison on open-source and closed-source models}
In this section, we compare the performance of closed-source models such as Google Gemini 2.5 Flash, OpenAI GPT4o, Anthropic Claude Sonnet 4 with a subset of popular open source models including Mistral~\cite{jiang2023mistral7b} which solved (0/65) main problems and (24/283) subproblems. Gemma3~\cite{gemma3} solved (1/65) main problems and (39/283) subproblems. In comparison, Llama 4~\cite{llama4} and Deepseek R1~\cite{deepseekr1} improved the performance compared to the other two with DeepSeek R1 solving (4/65) main problems and (84/283) subproblems, which doesn't even surpass the SciCode baseline with GPT 4o. As shown in Table~\ref{tab:abl_cl_op_model}, closed-source models consistently and substantially outperform their open-source counterparts across our scientific coding benchmarks. We attribute this performance gap to factors such as larger and more diverse training data, proprietary fine-tuning strategies, and stronger alignment mechanisms available to closed-source systems. An important direction for future research lies in domain specific fine-tuning of open-source models using curated scientific datasets, research papers and specialized coding repositories (e.g., GitHub projects). Such efforts could significantly improve alignment with scientific problem-solving and reduce the current performance gap with proprietary systems.


\begin{figure}[h!]
    \centering
    \includegraphics[width=1\linewidth]{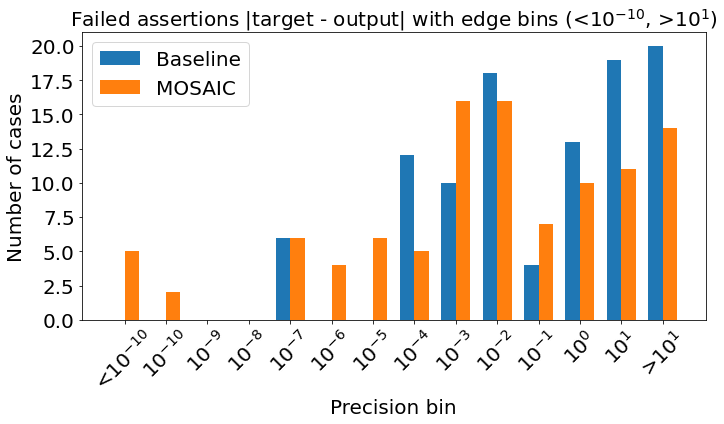}
    \caption{\textbf{Precision differences between target and generated outputs.} Compared to the baseline, MOSAIC produces a larger proportion of executable code, which introduces slightly more detectable errors. However, \textbf{MOSAIC outputs exhibit substantially smaller deviations} from the target values, indicating improved numerical precision.}
    \label{fig:precision_diff}
\end{figure}


\subsection{Understanding the performance numbers}
While Figure~\ref{fig:error_types} highlights the distribution of syntactic and semantic errors, it is also essential to evaluate the numerical precision of these assertion errors relative to the target values. To this end, we employ the SciCode test suite to systematically assess the outputs produced by MOSAIC. Figure~\ref{fig:precision_diff} reports the distribution of precision differences, categorized into bins ranging from $<10^{-10}$ to $>10^1$. The results indicate a reduction in large deviations, with MOSAIC consistently yielding outputs of higher precision. These findings highlight the framework’s effectiveness in improving both accuracy and precision. Importantly, this improvement is a direct consequence of MOSAIC’s student–teacher structure and its sequence of specialized agents, which encourage step-by-step reasoning, iterative self-correction, and targeted debugging. By guiding the model through intermediate rationales and consolidating context, MOSAIC generates algorithms that are not only more precise but also better aligned with the underlying problem structure. Such improvements provide a more informative diagnostic perspective than binary pass/fail outcomes, as they reveal partial correctness that would otherwise remain hidden. Moreover, the reduction in deviation is particularly significant in scientific and engineering contexts, where small numerical discrepancies can accumulate and substantially affect downstream analyses. Collectively, these results showcase the effectiveness of MOSAIC in enhancing both the accuracy and reliability of scientific code generation.


\section{Limitations and Future Work}
While our framework, MOSAIC, demonstrates the potential of designing and carefully orchestrating multi-agentic systems to benefit scientific coding and reasoning tasks, a few aspects remain beyond the scope of our initial investigation. One limitation of our work is that we use the domain split information provided in the SciCode dataset (split into physics, chemistry, biology, mathematics, and materials science). In our experiments, we tried domain bucketing using the LLM. We relied on the prior knowledge of the LLM and used a list of generic domain specific keywords. We observed that it often struggled to match problems to the correct domain, resulting in a performance drop of approximately 10–12\%. We have provided all the templates of the prompts used in the Appendix~\ref{sec:appendix} for reproducibility. Additionally, since MOSAIC is designed as a training free, LLM agnostic framework, we are bounded by the capabilities of the LLM backbone used in our agentic framework.

For future work, we plan to strengthen MOSAIC in several exciting directions. We will explore fine-tuning approaches that incorporate domain-specific knowledge from scientific datasets, libraries, and tools to improve accuracy in specialized areas such as physics, bioinformatics, and other computational sciences. In this study, we intentionally set aside such design choices to highlight the key drawbacks of existing baseline methods and to maximize performance in a training-free setup.

We also intend to develop heterogeneous agent configurations where different LLM backbones specialize in roles such as planning, code generation, debugging, and optimization, working collaboratively within the framework. We provided some initial results and insights on the strengths of different LLM backbones as different agents in Section~\ref{subsec:llm_backbone}. Another direction is reinforcement learning from execution feedback, enabling agents to refine solutions based on runtime performance and scalability rather than correctness alone. Together, these efforts will enhance MOSAIC’s adaptability and effectiveness for complex scientific challenges.

\section{Conclusion}
\label{sec:conclusion}
In this paper, we introduced MOSAIC, an intelligent multi-agent orchestration framework for tackling complex scientific coding challenges. Drawing inspiration from knowledge distillation, MOSAIC employs a teacher–student paradigm in which the teacher guides the student through few-shot prompting and self-reflection to generate robust pseudocode. The framework mitigates hallucinations via a consolidated context window and performs iterative debugging without the need for sample test sets, thereby achieving substantial improvements over baseline performance. Extensive experiments and ablation studies validate the effectiveness and robustness of the proposed approach. 

\section*{Acknowledgement}
This research was supported by the U.S. Department of Energy, Office of Science, Advanced Scientific Computing
Computing Research, through the SciDAC-RAPIDS2 institute under Contract DE-AC02-06CH11357. 

\bibliographystyle{IEEEtran}
\bibliography{bibliography}

\begin{thebibliography}{10}
\providecommand{\url}[1]{#1}
\csname url@samestyle\endcsname
\providecommand{\newblock}{\relax}
\providecommand{\bibinfo}[2]{#2}
\providecommand{\BIBentrySTDinterwordspacing}{\spaceskip=0pt\relax}
\providecommand{\BIBentryALTinterwordstretchfactor}{4}
\providecommand{\BIBentryALTinterwordspacing}{\spaceskip=\fontdimen2\font plus
\BIBentryALTinterwordstretchfactor\fontdimen3\font minus \fontdimen4\font\relax}
\providecommand{\BIBforeignlanguage}[2]{{%
\expandafter\ifx\csname l@#1\endcsname\relax
\typeout{** WARNING: IEEEtran.bst: No hyphenation pattern has been}%
\typeout{** loaded for the language `#1'. Using the pattern for}%
\typeout{** the default language instead.}%
\else
\language=\csname l@#1\endcsname
\fi
#2}}
\providecommand{\BIBdecl}{\relax}
\BIBdecl

\bibitem{chen2021evaluatinglargelanguagemodels}
\BIBentryALTinterwordspacing
M.~Chen, J.~Tworek, H.~Jun, Q.~Yuan, H.~P.~O. Pinto, J.~Kaplan, H.~Edwards, Y.~Burda, N.~Joseph, G.~Brockman \emph{et~al.}, ``Evaluating large language models trained on code,'' \emph{arXiv preprint arXiv:2107.03374}, 2021. [Online]. Available: \url{https://arxiv.org/abs/2107.03374}
\BIBentrySTDinterwordspacing

\bibitem{cot}
J.~Wei, X.~Wang, D.~Schuurmans, M.~Bosma, F.~Xia, E.~Chi, Q.~V. Le, D.~Zhou \emph{et~al.}, ``Chain-of-thought prompting elicits reasoning in large language models,'' in \emph{Advances in Neural Information Processing Systems}, vol.~35, 2022, pp. 24\,824--24\,837.

\bibitem{chen2022codet}
B.~Chen, F.~Zhang, A.~Nguyen, D.~Zan, Z.~Lin, J.-G. Lou, and W.~Chen, ``Codet: Code generation with generated tests,'' \emph{arXiv preprint arXiv:2207.10397}, 2022.

\bibitem{parvez2021retrieval}
M.~R. Parvez, W.~U. Ahmad, S.~Chakraborty, B.~Ray, and K.-W. Chang, ``Retrieval augmented code generation and summarization,'' \emph{arXiv preprint arXiv:2108.11601}, 2021.

\bibitem{shum-etal-2023-automatic}
K.~Shum, S.~Diao, and T.~Zhang, ``Automatic prompt augmentation and selection with chain-of-thought from labeled data,'' in \emph{Findings of the Association for Computational Linguistics: EMNLP 2023}, Singapore, December 2023, pp. 12\,113--12\,139.

\bibitem{zhang2022automatic}
Z.~Zhang, A.~Zhang, M.~Li, and A.~Smola, ``Automatic chain of thought prompting in large language models,'' \emph{arXiv preprint arXiv:2210.03493}, 2022.

\bibitem{zhou2023language}
A.~Zhou, K.~Yan, M.~Shlapentokh-Rothman, H.~Wang, and Y.-X. Wang, ``Language agent tree search unifies reasoning acting and planning in language models,'' \emph{arXiv preprint arXiv:2310.04406}, 2023.

\bibitem{yasunaga2023large}
M.~Yasunaga, X.~Chen, Y.~Li, P.~Pasupat, J.~Leskovec, P.~Liang, E.~Chi, and D.~Zhou, ``Large language models as analogical reasoners,'' \emph{arXiv preprint arXiv:2310.01714}, 2023.

\bibitem{agentcoder}
\BIBentryALTinterwordspacing
D.~Huang, J.~M. Zhang, M.~Luck, Q.~Bu, Y.~Qing, and H.~Cui, ``Agentcoder: Multi-agent-based code generation with iterative testing and optimisation,'' \emph{arXiv preprint arXiv:2312.13010}, 2024. [Online]. Available: \url{https://arxiv.org/abs/2312.13010}
\BIBentrySTDinterwordspacing

\bibitem{mapcoder}
\BIBentryALTinterwordspacing
M.~A. Islam, M.~E. Ali, and M.~R. Parvez, ``Mapcoder: Multi-agent code generation for competitive problem solving,'' \emph{arXiv preprint arXiv:2405.11403}, 2024. [Online]. Available: \url{https://arxiv.org/abs/2405.11403}
\BIBentrySTDinterwordspacing

\bibitem{codesim}
\BIBentryALTinterwordspacing
------, ``Codesim: Multi-agent code generation and problem solving through simulation-driven planning and debugging,'' \emph{arXiv preprint arXiv:2502.05664}, 2025. [Online]. Available: \url{https://arxiv.org/abs/2502.05664}
\BIBentrySTDinterwordspacing

\bibitem{mallick2024chatvis}
T.~Mallick, O.~Yildiz, D.~Lenz, and T.~Peterka, ``Chatvis: Automating scientific visualization with a large language model,'' in \emph{Proc. SC24-W: Workshops of the International Conference for High Performance Computing, Networking, Storage and Analysis}.\hskip 1em plus 0.5em minus 0.4em\relax IEEE, 2024, pp. 49--55.

\bibitem{LLM_hallucinate}
Z.~Zhang, C.~Wang, Y.~Wang, E.~Shi, Y.~Ma, W.~Zhong, J.~Chen, M.~Mao, and Z.~Zheng, ``Llm hallucinations in practical code generation: Phenomena, mechanism, and mitigation,'' \emph{Proc. ACM Softw. Eng.}, vol.~2, no. ISSTA, pp. 1--23, June 2025.

\bibitem{Wang2023ScientificDI}
H.~Wang, T.~Fu, Y.~Du, W.~Gao, K.~Huang, Z.~Liu, P.~Chandak, S.~Liu, P.~Van~Katwyk, A.~Deac, A.~Anandkumar, K.~J. Bergen, C.~P. Gomes, S.~Ho, P.~Kohli, J.~Lasenby, J.~Leskovec, T.-Y. Liu, A.~K. Manrai, D.~S. Marks, B.~Ramsundar, L.~Song, J.~Sun, J.~Tang, P.~Velickovic, M.~Welling, L.~Zhang, C.~W. Coley, Y.~Bengio, and M.~Zitnik, ``Scientific discovery in the age of artificial intelligence,'' \emph{Nature}, vol. 620, pp. 47--60, 2023.

\bibitem{ai4science2023impactlargelanguagemodels}
\BIBentryALTinterwordspacing
{Microsoft Research AI4Science} and {Microsoft Azure Quantum}, ``The impact of large language models on scientific discovery: a preliminary study using gpt-4,'' \emph{arXiv preprint arXiv:2311.07361}, 2023. [Online]. Available: \url{https://arxiv.org/abs/2311.07361}
\BIBentrySTDinterwordspacing

\bibitem{zhang2024scientificllms_survey}
Y.~Zhang \emph{et~al.}, ``A comprehensive survey of scientific large language models and their applications in scientific discovery,'' \emph{arXiv preprint}, 2024.

\bibitem{taylor2022galactica}
P.~Taylor, J.~Lucas, E.~Zhai \emph{et~al.}, ``Galactica: A large language model for science,'' \emph{arXiv preprint arXiv:2211.09085}, 2022.

\bibitem{biogpt}
R.~Luo, L.~Sun, Y.~Xia, T.~Qin, S.~Zhang, H.~Poon, and T.-Y. Liu, ``Biogpt: generative pre-trained transformer for biomedical text generation and mining,'' \emph{Briefings in Bioinformatics}, vol.~23, no.~6, pp. 1--9, September 2022.

\bibitem{edwards2024sciem}
\BIBentryALTinterwordspacing
C.~Edwards, J.~Chen, and M.~Johnson, ``Sciem: A multimodal benchmark for scientific entity matching,'' \emph{arXiv preprint arXiv:2403.01234}, 2024. [Online]. Available: \url{https://arxiv.org/abs/2403.01234}
\BIBentrySTDinterwordspacing

\bibitem{li2024matbenchllm}
\BIBentryALTinterwordspacing
Y.~Li, H.~Zhang, and K.~Wu, ``Matbench-llm: Evaluating large language models on materials science benchmarks,'' \emph{arXiv preprint arXiv:2404.05678}, 2024. [Online]. Available: \url{https://arxiv.org/abs/2404.05678}
\BIBentrySTDinterwordspacing

\bibitem{wang2023selfconsistencyimproveschainthought}
\BIBentryALTinterwordspacing
X.~Wang, J.~Wei, D.~Schuurmans, Q.~Le, E.~Chi, S.~Narang, A.~Chowdhery, and D.~Zhou, ``Self-consistency improves chain of thought reasoning in language models,'' \emph{arXiv preprint arXiv:2203.11171}, 2023. [Online]. Available: \url{https://arxiv.org/abs/2203.11171}
\BIBentrySTDinterwordspacing

\bibitem{selfrefine}
\BIBentryALTinterwordspacing
A.~Madaan \emph{et~al.}, ``Self-refine: Iterative refinement with self-feedback,'' in \emph{Proc. ICML}, 2023. [Online]. Available: \url{https://arxiv.org/abs/2303.17651}
\BIBentrySTDinterwordspacing

\bibitem{treeofthoughts}
\BIBentryALTinterwordspacing
S.~Yao \emph{et~al.}, ``Tree of thoughts: Deliberate problem solving with large language models,'' \emph{arXiv preprint arXiv:2305.10601}, 2023. [Online]. Available: \url{https://arxiv.org/abs/2305.10601}
\BIBentrySTDinterwordspacing

\bibitem{graphoffthoughts}
\BIBentryALTinterwordspacing
M.~Besta \emph{et~al.}, ``Graph of thoughts: Solving elaborate problems with large language models,'' \emph{arXiv preprint arXiv:2308.09687}, 2023. [Online]. Available: \url{https://arxiv.org/abs/2308.09687}
\BIBentrySTDinterwordspacing

\bibitem{reflexion}
\BIBentryALTinterwordspacing
N.~Shinn \emph{et~al.}, ``Reflexion: Language agents with verbal reinforcement learning,'' in \emph{Proc. NeurIPS}, 2023. [Online]. Available: \url{https://arxiv.org/abs/2303.11366}
\BIBentrySTDinterwordspacing

\bibitem{jin2019pubmedqa}
\BIBentryALTinterwordspacing
Q.~Jin \emph{et~al.}, ``Pubmedqa: A dataset for biomedical research question answering,'' in \emph{Proc. EMNLP}, 2019, pp. 2567--2577. [Online]. Available: \url{https://aclanthology.org/D19-1259/}
\BIBentrySTDinterwordspacing

\bibitem{jin2021medqa}
D.~Jin, E.~Pan, N.~Oufattole, W.-H. Weng, H.~Fang, and P.~Szolovits, ``What disease does this patient have? a large-scale open domain question answering dataset from medical exams,'' \emph{arXiv preprint arXiv:2009.13081}, 2020.

\bibitem{lu2022scienceqa}
Z.~Lu, Y.~Wei, H.~Tang \emph{et~al.}, ``Scienceqa: Towards standardized dataset for science question answering,'' in \emph{Proc. EMNLP}, 2022.

\bibitem{rodriguez2023benchflow}
M.~Rodriguez, P.~Singh, and X.~Chen, ``Benchflow: Benchmarking llm-driven scientific workflows,'' \emph{Journal of Computational Science}, vol.~52, p. 101507, 2023.

\bibitem{face4rag}
\BIBentryALTinterwordspacing
Y.~Zhang \emph{et~al.}, ``Face4rag: A factual consistency evaluation benchmark for retrieval-augmented generation,'' \emph{arXiv preprint arXiv:2407.01080}, 2024. [Online]. Available: \url{https://arxiv.org/abs/2407.01080}
\BIBentrySTDinterwordspacing

\bibitem{scicode}
\BIBentryALTinterwordspacing
M.~Tian, L.~Gao, S.~D. Zhang, X.~Chen, C.~Fan, X.~Guo, R.~Haas, P.~Ji, K.~Krongchon, Y.~Li, S.~Liu, D.~Luo, Y.~Ma, H.~Tong, K.~Trinh, C.~Tian, Z.~Wang, B.~Wu, Y.~Xiong, S.~Yin, M.~Zhu, K.~Lieret, Y.~Lu, G.~Liu, Y.~Du, T.~Tao, O.~Press, J.~Callan, E.~Huerta, and H.~Peng, ``Scicode: A research coding benchmark curated by scientists,'' \emph{arXiv preprint arXiv:2407.13168}, 2024. [Online]. Available: \url{https://arxiv.org/abs/2407.13168}
\BIBentrySTDinterwordspacing

\bibitem{chen2021evaluating}
M.~Chen, J.~Tworek, H.~Jun \emph{et~al.}, ``Evaluating large language models trained on code,'' \emph{arXiv preprint arXiv:2107.03374}, 2021.

\bibitem{codegen}
\BIBentryALTinterwordspacing
E.~Nijkamp, B.~Pang, H.~Hayashi, L.~Tu, H.~Wang, Y.~Zhou, S.~Savarese, and C.~Xiong, ``Codegen: An open large language model for code with multi-turn program synthesis,'' \emph{arXiv preprint arXiv:2203.13474}, 2023. [Online]. Available: \url{https://arxiv.org/abs/2203.13474}
\BIBentrySTDinterwordspacing

\bibitem{santacoder}
\BIBentryALTinterwordspacing
L.~Ben~Allal, R.~Li, D.~Kocetkov, C.~Mou, C.~Akiki, C.~M. Ferrandis, N.~Muennighoff, M.~Mishra, A.~Gu, M.~Dey \emph{et~al.}, ``Santacoder: Don't reach for the stars!'' \emph{arXiv preprint arXiv:2301.03988}, 2023. [Online]. Available: \url{https://arxiv.org/abs/2301.03988}
\BIBentrySTDinterwordspacing

\bibitem{starcoder}
\BIBentryALTinterwordspacing
R.~Li, L.~Ben~Allal, Y.~Zi, N.~Muennighoff, D.~Kocetkov, C.~Mou, M.~Marone, C.~Akiki, J.~Li, J.~Chim \emph{et~al.}, ``Starcoder: May the source be with you!'' \emph{arXiv preprint arXiv:2305.06161}, 2023. [Online]. Available: \url{https://arxiv.org/abs/2305.06161}
\BIBentrySTDinterwordspacing

\bibitem{APPS}
D.~Hendrycks, S.~Basart, S.~Kadavath, M.~Mazeika, A.~Arora, E.~Guo, C.~Burns, S.~Puranik, H.~He, D.~Song, and J.~Steinhardt, ``Measuring coding challenge competence with apps,'' in \emph{NeurIPS}, 2021.

\bibitem{MBPP}
\BIBentryALTinterwordspacing
J.~Austin, A.~Odena, M.~I. Nye, M.~Bosma, H.~Michalewski, D.~Dohan, E.~Jiang, C.~Cai, M.~Terry, Q.~V. Le, and C.~Sutton, ``Program synthesis with large language models,'' \emph{CoRR}, vol. abs/2108.07732, 2021. [Online]. Available: \url{https://arxiv.org/abs/2108.07732}
\BIBentrySTDinterwordspacing

\bibitem{swebench}
\BIBentryALTinterwordspacing
C.~Jimenez \emph{et~al.}, ``Swe-bench: Can language models resolve real-world github issues?'' in \emph{Proc. ICLR}, 2024. [Online]. Available: \url{https://arxiv.org/abs/2310.06770}
\BIBentrySTDinterwordspacing

\bibitem{sweagent}
\BIBentryALTinterwordspacing
J.~Yang \emph{et~al.}, ``Swe-agent: Agent–computer interfaces enable automated software engineering,'' in \emph{Proc. NeurIPS}, 2024. [Online]. Available: \url{https://arxiv.org/abs/2405.15793}
\BIBentrySTDinterwordspacing

\bibitem{mathchat}
Y.~Wu \emph{et~al.}, ``Mathchat: Converse to tackle challenging math problems with llm agents,'' \emph{arXiv preprint}, 2024.

\bibitem{toolformer}
\BIBentryALTinterwordspacing
T.~Schick \emph{et~al.}, ``Toolformer: Language models can teach themselves to use tools,'' \emph{arXiv preprint arXiv:2302.04761}, 2023. [Online]. Available: \url{https://arxiv.org/abs/2302.04761}
\BIBentrySTDinterwordspacing

\bibitem{autogen}
\BIBentryALTinterwordspacing
Q.~Wu \emph{et~al.}, ``Autogen: Enabling next-gen llm applications via multi-agent conversation,'' \emph{arXiv preprint arXiv:2308.08155}, 2023. [Online]. Available: \url{https://arxiv.org/abs/2308.08155}
\BIBentrySTDinterwordspacing

\bibitem{hinton2015distillingknowledgeneuralnetwork}
\BIBentryALTinterwordspacing
G.~Hinton, O.~Vinyals, and J.~Dean, ``Distilling the knowledge in a neural network,'' \emph{arXiv preprint arXiv:1503.02531}, 2015. [Online]. Available: \url{https://arxiv.org/abs/1503.02531}
\BIBentrySTDinterwordspacing

\bibitem{banerjee2024llmshallucinateneedlive}
\BIBentryALTinterwordspacing
S.~Banerjee, A.~Agarwal, and S.~Singla, ``Llms will always hallucinate, and we need to live with this,'' \emph{arXiv preprint arXiv:2409.05746}, 2024. [Online]. Available: \url{https://arxiv.org/abs/2409.05746}
\BIBentrySTDinterwordspacing

\bibitem{humaneval}
M.~Chen, J.~Tworek, H.~Jun, Q.~Yuan, H.~P.~O. Pinto, J.~Kaplan, H.~Edwards, Y.~Burda, N.~Joseph, G.~Brockman \emph{et~al.}, ``Evaluating large language models trained on code,'' \emph{arXiv preprint arXiv:2107.03374}, 2021.

\bibitem{xu2024llmsabstractionreasoningcorpus}
\BIBentryALTinterwordspacing
Y.~Xu, W.~Li, P.~Vaezipoor, S.~Sanner, and E.~B. Khalil, ``Llms and the abstraction and reasoning corpus: Successes, failures, and the importance of object-based representations,'' \emph{arXiv preprint arXiv:2305.18354}, 2024. [Online]. Available: \url{https://arxiv.org/abs/2305.18354}
\BIBentrySTDinterwordspacing

\bibitem{jiang2023mistral7b}
\BIBentryALTinterwordspacing
A.~Q. Jiang, A.~Sablayrolles, A.~Mensch, C.~Bamford, D.~S. Chaplot, D.~de~las Casas, F.~Bressand, G.~Lengyel, G.~Lample, L.~Saulnier \emph{et~al.}, ``Mistral 7b,'' \emph{arXiv preprint arXiv:2310.06825}, 2023. [Online]. Available: \url{https://arxiv.org/abs/2310.06825}
\BIBentrySTDinterwordspacing

\bibitem{gemma3}
\BIBentryALTinterwordspacing
{Gemma Team}, A.~Kamath, J.~Ferret, S.~Pathak, N.~Vieillard, R.~Merhej, S.~Perrin, T.~Matejovicova, A.~Ramé, M.~Rivière \emph{et~al.}, ``Gemma 3 technical report,'' \emph{arXiv preprint arXiv:2503.19786}, 2025. [Online]. Available: \url{https://arxiv.org/abs/2503.19786}
\BIBentrySTDinterwordspacing

\bibitem{llama4}
\BIBentryALTinterwordspacing
H.~Touvron, T.~Lavril, G.~Izacard, X.~Martinet, M.-A. Lachaux, T.~Lacroix, B.~Rozière, N.~Goyal, E.~Hambro, F.~Azhar, A.~Rodriguez, A.~Joulin, E.~Grave, and G.~Lample, ``Llama: Open and efficient foundation language models,'' \emph{arXiv preprint arXiv:2302.13971}, 2023. [Online]. Available: \url{https://arxiv.org/abs/2302.13971}
\BIBentrySTDinterwordspacing

\bibitem{deepseekr1}
\BIBentryALTinterwordspacing
{DeepSeek-AI}, D.~Guo, D.~Yang, H.~Zhang, J.~Song, R.~Zhang, R.~Xu, Q.~Zhu, S.~Ma, P.~Wang, X.~Bi \emph{et~al.}, ``Deepseek-r1: Incentivizing reasoning capability in llms via reinforcement learning,'' \emph{arXiv preprint arXiv:2501.12948}, 2025. [Online]. Available: \url{https://arxiv.org/abs/2501.12948}
\BIBentrySTDinterwordspacing

\end{thebibliography}
\clearpage
\appendix

\section{Appendix}
\label{sec:appendix}

\begin{figure*}[ht!]
    \centering
    \includegraphics[width=1\linewidth]{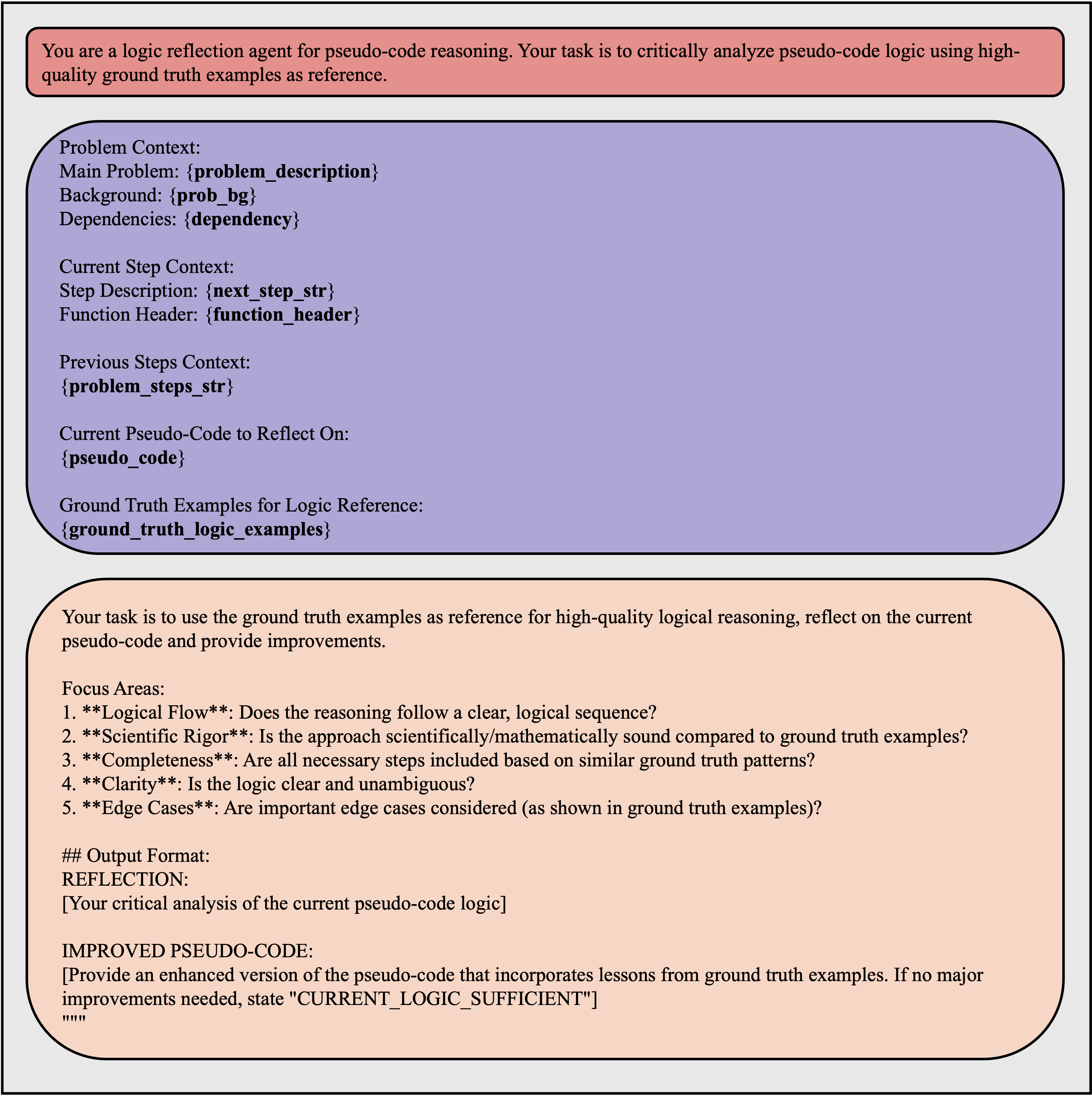}
    \caption{The prompt template for our \textbf{Self-Reflection Agent}, which analyzes the ground truth code from the validation set to understand patterns from each domain. And prepare gold standard domain specific pseudocode to use as few shot examples by the Rationale Agent}
    \label{appendix_fig:reflect-prompt}
\end{figure*}

\begin{figure*}[ht!]
    \centering
    \includegraphics[width=1\linewidth]{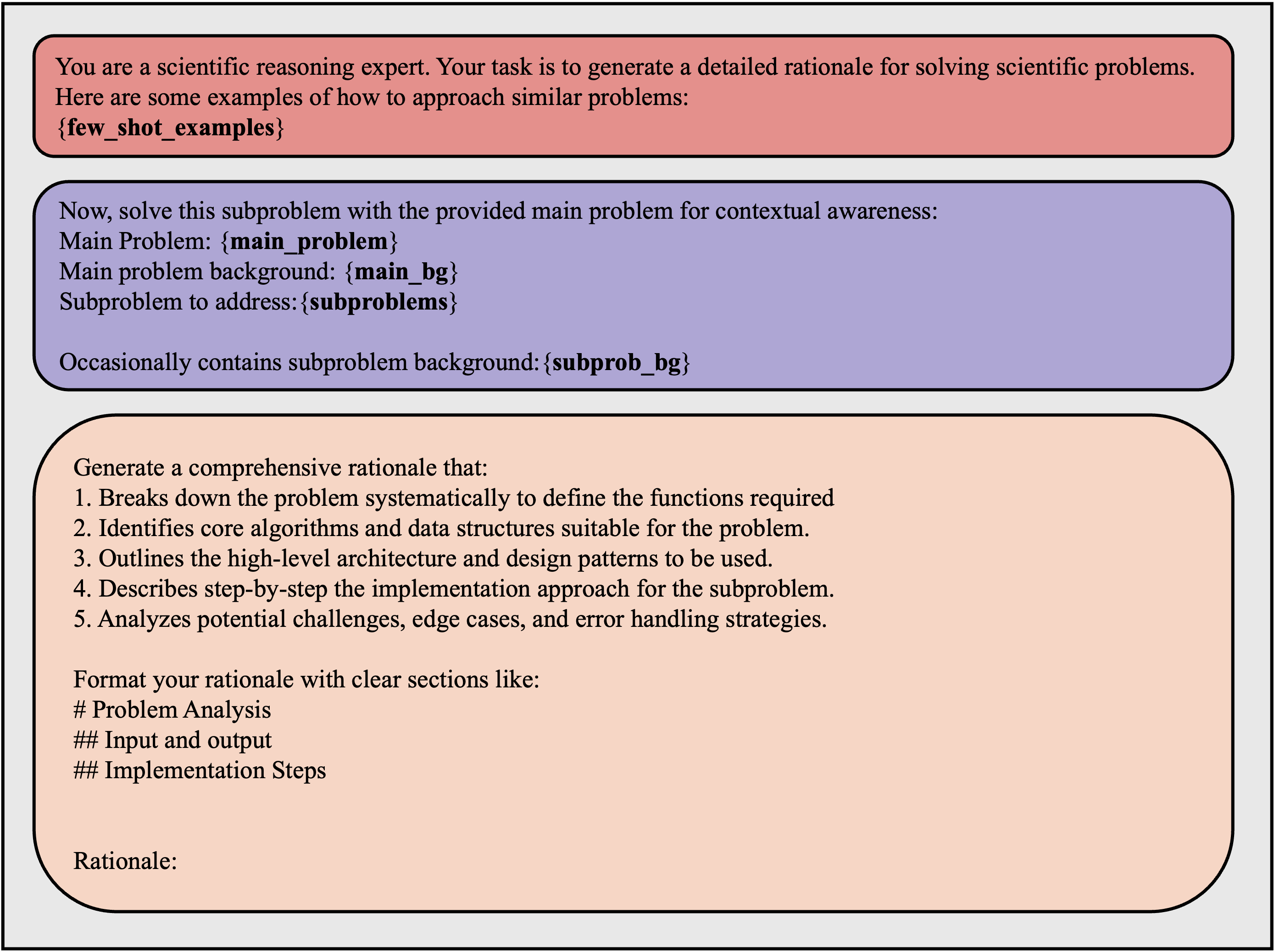}
    \caption{The prompt template for our \textbf{Rationale Agent}. Guided by the few-shot examples from the self-reflection agent it converts the subproblem at hand into a detailed rationale than can be used by the coding agent to convert the algorithm into working code.}
    \label{appendix_fig:rationale-prompt}
\end{figure*}

\begin{figure*}[h]
    \centering
    \includegraphics[width=1\linewidth]{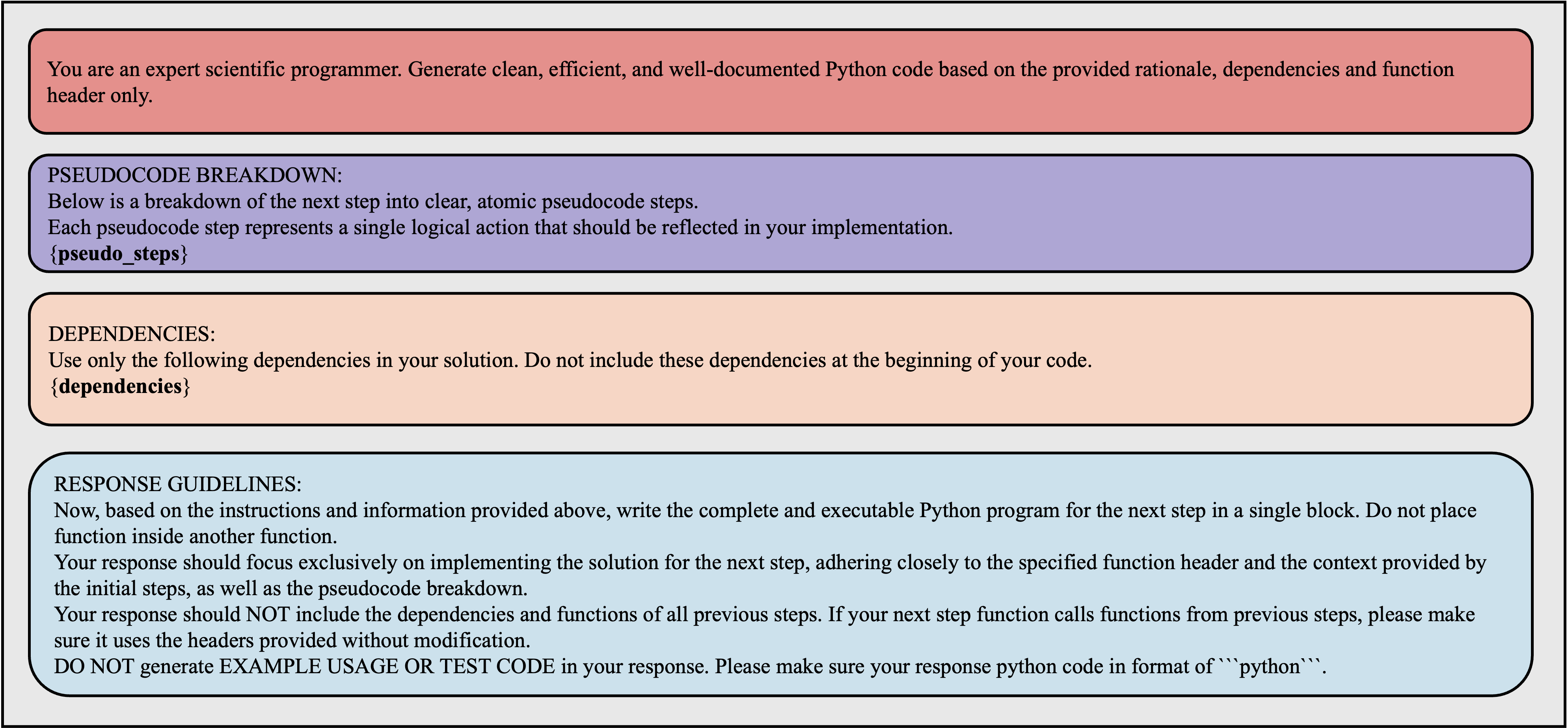}
    \caption{The prompt template for our \textbf{Coding Agent}. Once the context is set for the agent, it is guided to convert each step of the pseudocode into code.}
    \label{appendix_fig:code-prompt}
\end{figure*}

\begin{figure*}[ht!]
    \centering
    \includegraphics[width=1\linewidth]{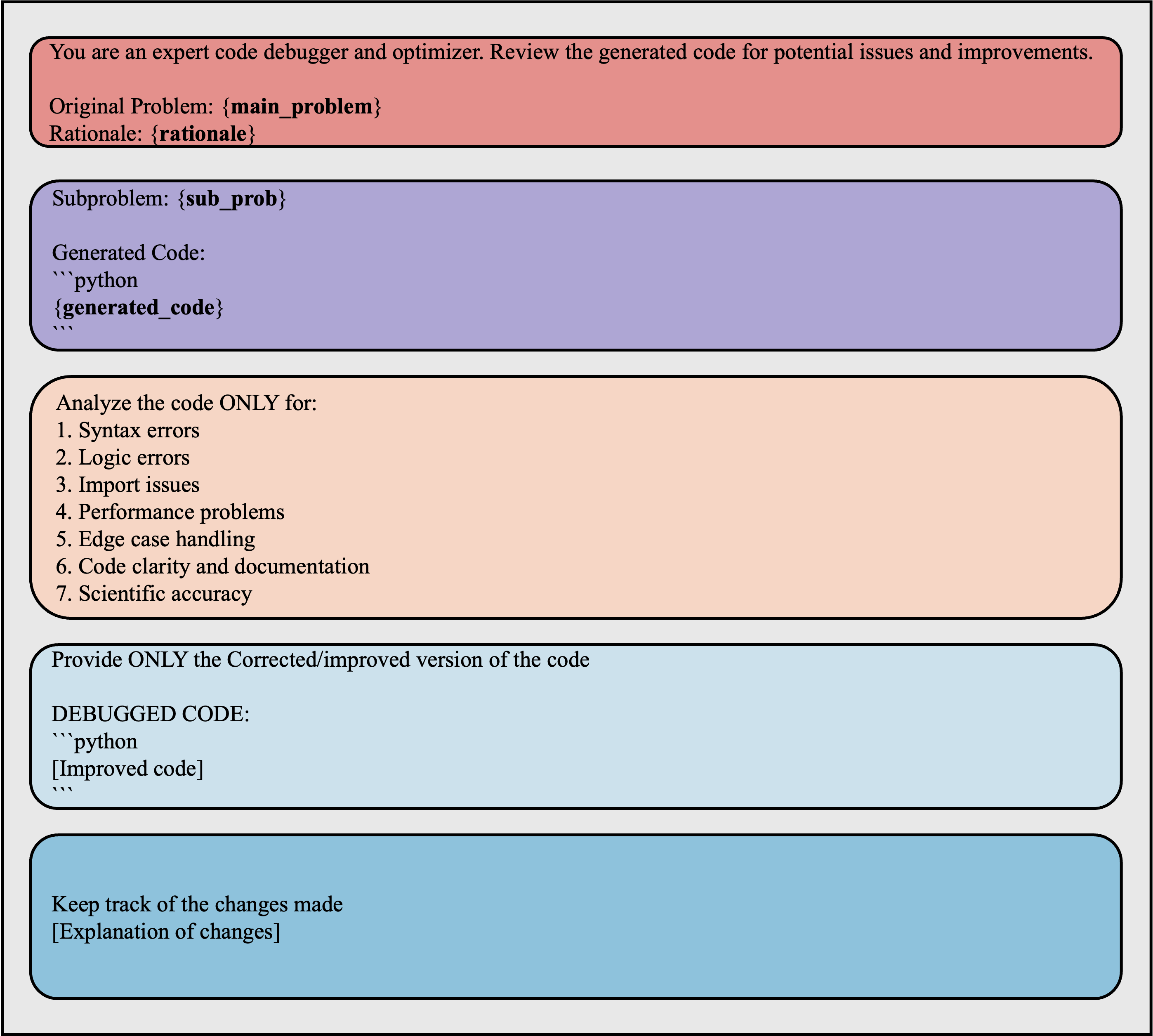}
    \caption{The prompt template for our \textbf{Debugger Agent}, which focuses on identifying and correcting syntactic errors to ensure executability of the code generated by the Coding Agent.}
    \label{appendix_fig:debug-prompt}
\end{figure*}

\begin{figure*}[ht!]
    \centering
    \includegraphics[width=1\linewidth]{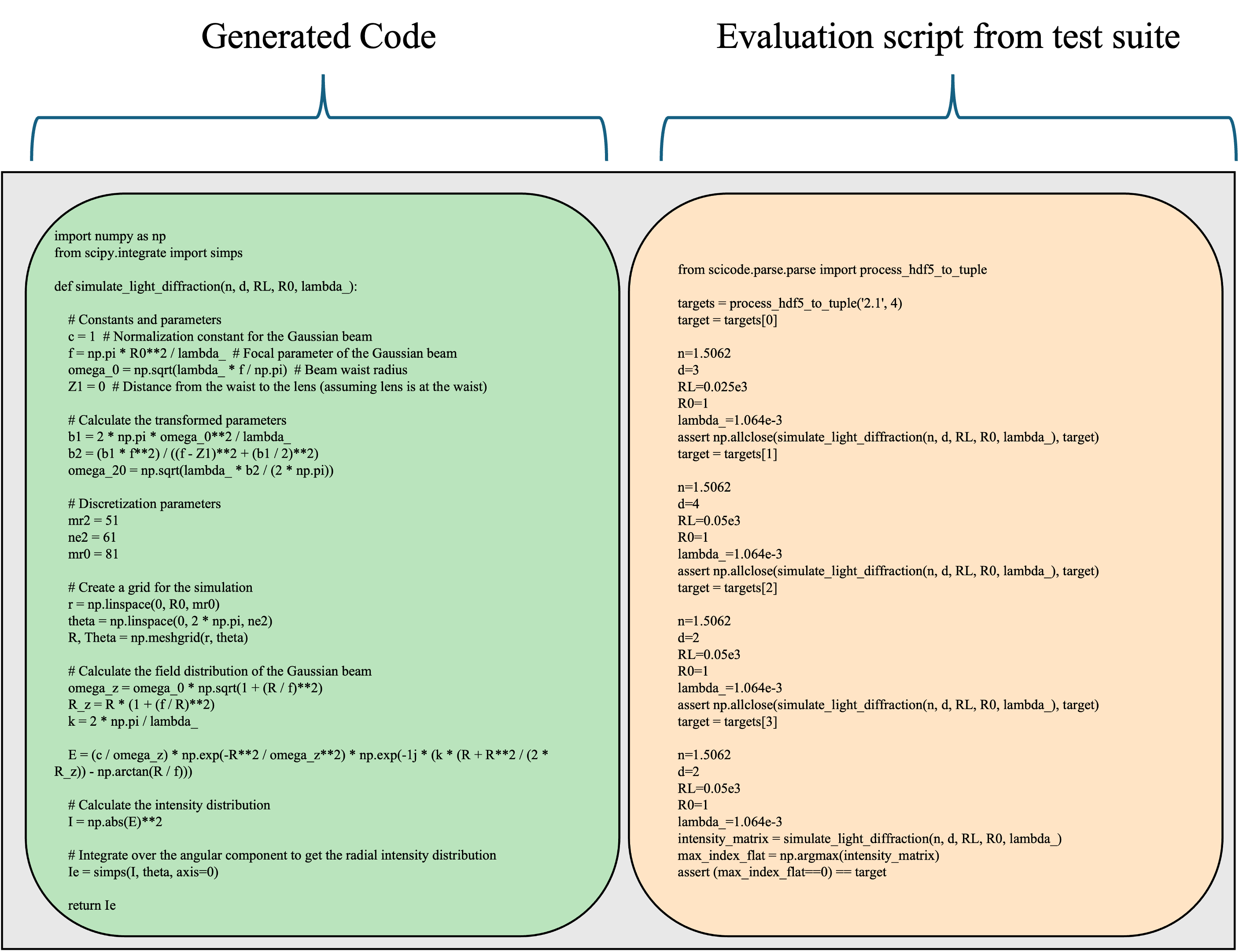}
    \caption{A sample of code generated by MOSAIC, alongside the evaluation code provided by the test suite to verify correctness.}
    \label{appendix_fig:code-generated}
\end{figure*}

In our appendix, we provide the full set of prompt templates used in our MOSAIC framework. As described in the method section~\ref{sec:method} of the main paper, our framework consists of 4 core agents (Self Reflection, Rationale, Coding and Debugger) in a student teacher setup inspired by knowledge distillation. We showcase the prompt templates of these agents in the below Figures~\ref{appendix_fig:reflect-prompt}, \ref{appendix_fig:rationale-prompt}, \ref{appendix_fig:code-prompt} and \ref{appendix_fig:debug-prompt}. We also introduce and implement a Consolidated Context Window (CCW) mechanism to minimize hallucinations and improve the quality of generated code. Together, these prompt mechanism define the workflow proposed in the main paper. We also include the output of a subproblem code generated by MOSAIC along with the evaluation script from the test suite provided by SciCode~\cite{scicode}, in Figure~\ref{appendix_fig:code-generated}. All resources, including source code and prompt templates, will be released upon acceptance of this work.

\end{document}